\PassOptionsToPackage{prologue,dvipsnames}{xcolor}
\documentclass[10pt,twocolumn,letterpaper]{article}

\usepackage{cvpr}              % To produce the CAMERA-READY version
\usepackage[colorinlistoftodos,prependcaption,textsize=small]{todonotes}
\usepackage[dvipsnames]{xcolor}
\definecolor{cvprblue}{rgb}{0.21,0.49,0.74}
\usepackage[pagebackref,breaklinks,colorlinks,citecolor=cvprblue]{hyperref}
\usepackage[accsupp]{axessibility}

\title{Audio-Visual Speech Representation Expert for Enhanced \\Talking Face Video Generation and Evaluation}

\author{Dogucan Yaman\textsuperscript{1} \qquad Fevziye Irem Eyiokur\textsuperscript{1} \qquad Leonard Bärmann\textsuperscript{1} \qquad Seymanur Aktı\textsuperscript{1} \\ Hazım Kemal Ekenel\textsuperscript{2} \qquad Alexander Waibel\textsuperscript{1,3} \\
\textsuperscript{1}Karlsruhe Institute of Technology, \textsuperscript{2}Istanbul Technical University, \textsuperscript{3}Carnegie Mellon University\\
{\tt\small dogucan.yaman@kit.edu}}

\newcommand{\paragraphHeading}[1]{\vspace{0.1cm}\noindent\textbf{#1}}

\begin{document}
\maketitle
\begin{abstract}

In the task of talking face generation, the objective is to generate a face video with lips synchronized to the corresponding audio while preserving visual details and identity information. Current methods face the challenge of learning accurate lip synchronization while avoiding detrimental effects on visual quality, as well as robustly evaluating such synchronization. To tackle these problems, we propose utilizing an audio-visual speech representation expert (AV-HuBERT) for calculating lip synchronization loss during training. Moreover, leveraging AV-HuBERT's features, we introduce three novel lip synchronization evaluation metrics, aiming to provide a comprehensive assessment of lip synchronization performance. Experimental results, along with a detailed ablation study, demonstrate the effectiveness of our approach and the utility of the proposed evaluation metrics.

\end{abstract}    
\section{Introduction}
\label{sec:intro}

The goal of talking face generation is to create a video based on provided face and audio sequences, seeking synchronized lip movements that match the given audio while maintaining the identity and visual details. 
This task has gained considerable interest recently for its diverse applications, such as face dubbing and enhancement in video conferencing tools, film dubbing, and content creation~\cite{zhan2021multimodal,zhen2023human}.

\begin{figure}[tb]
  \centering
  \begin{subfigure}{0.49\linewidth}
    \includegraphics[width=\linewidth]{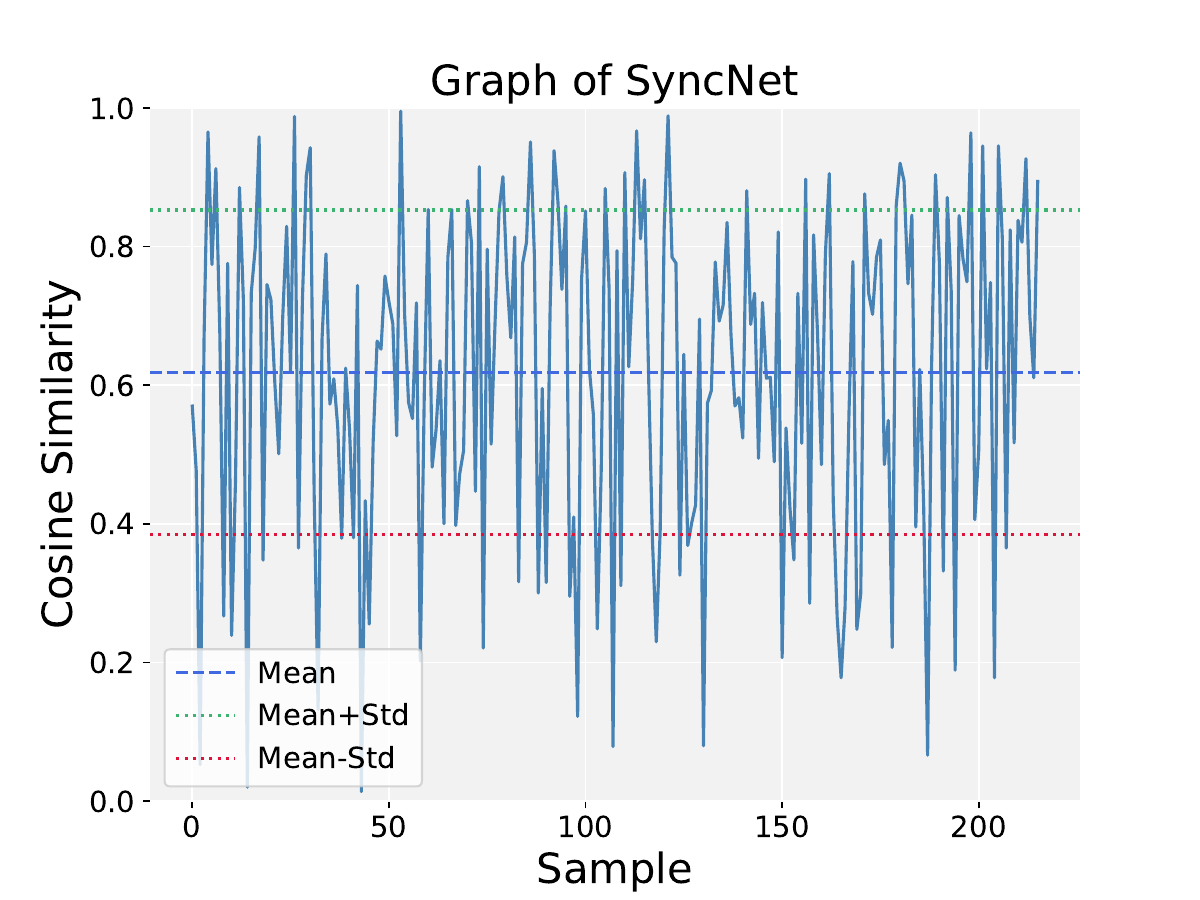}
    \caption{Cosine similarity}
    \label{fig:cosine_syncnet}
  \end{subfigure}
  \hfill
  \begin{subfigure}{0.49\linewidth}
    \includegraphics[trim={0cm 0cm 0cm 0cm},clip,width=\linewidth]{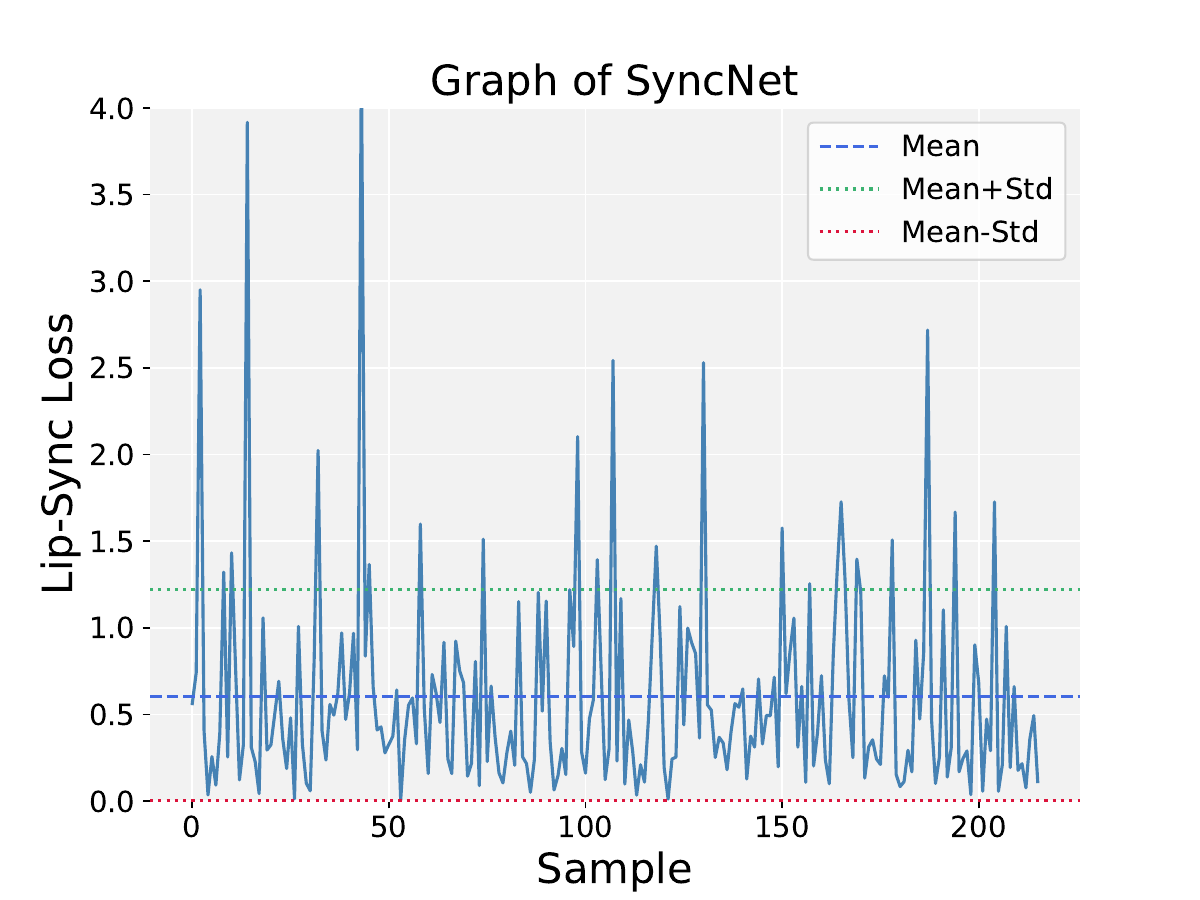}
    \caption{Lip-sync loss}
    \label{fig:logloss_syncnet}
  \end{subfigure}
  \begin{subfigure}{0.49\linewidth}
    \includegraphics[width=\linewidth]{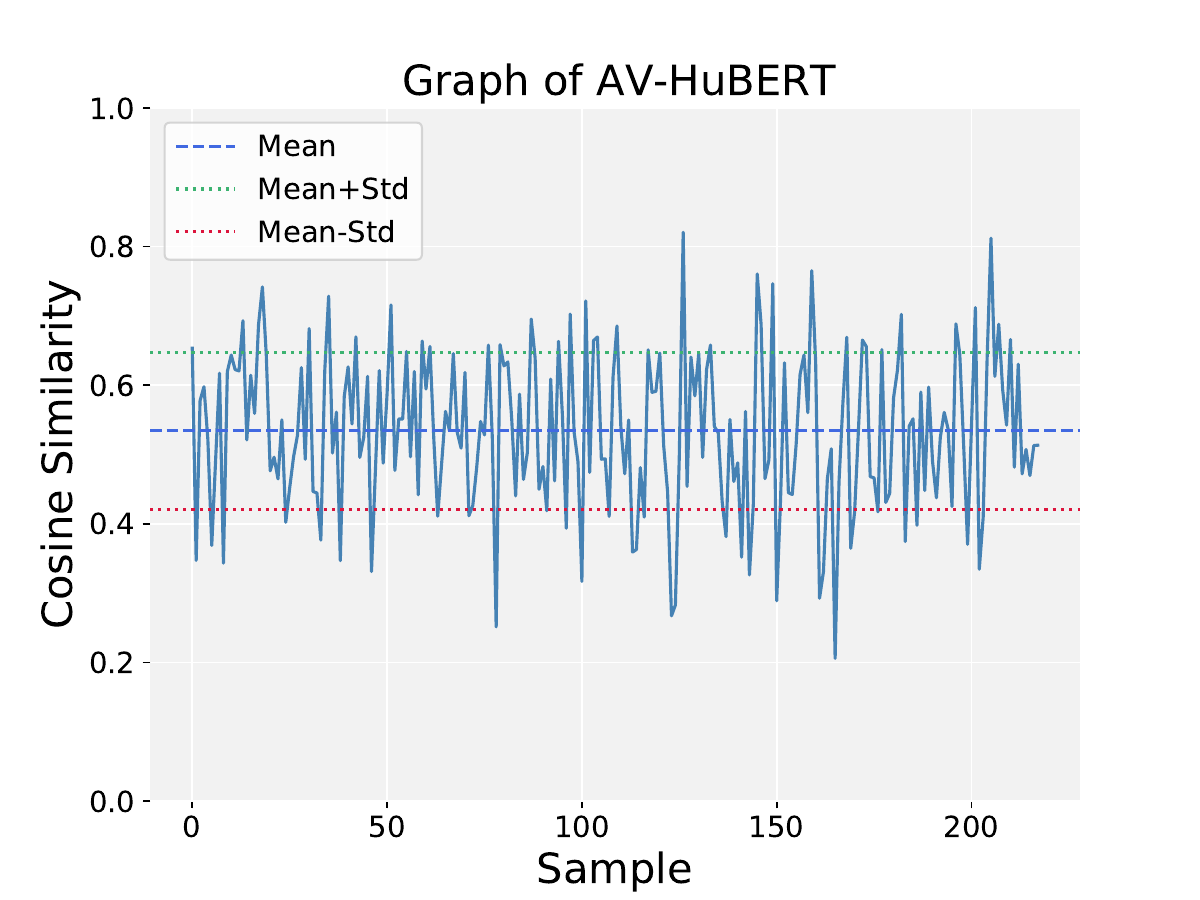}
    \caption{Cosine similarity}
    \label{fig:cosine_sim_avhubert}
  \end{subfigure}
  \hfill
  \begin{subfigure}{0.49\linewidth}
    \includegraphics[trim={0cm 0cm 0cm 0cm},clip,width=\linewidth]{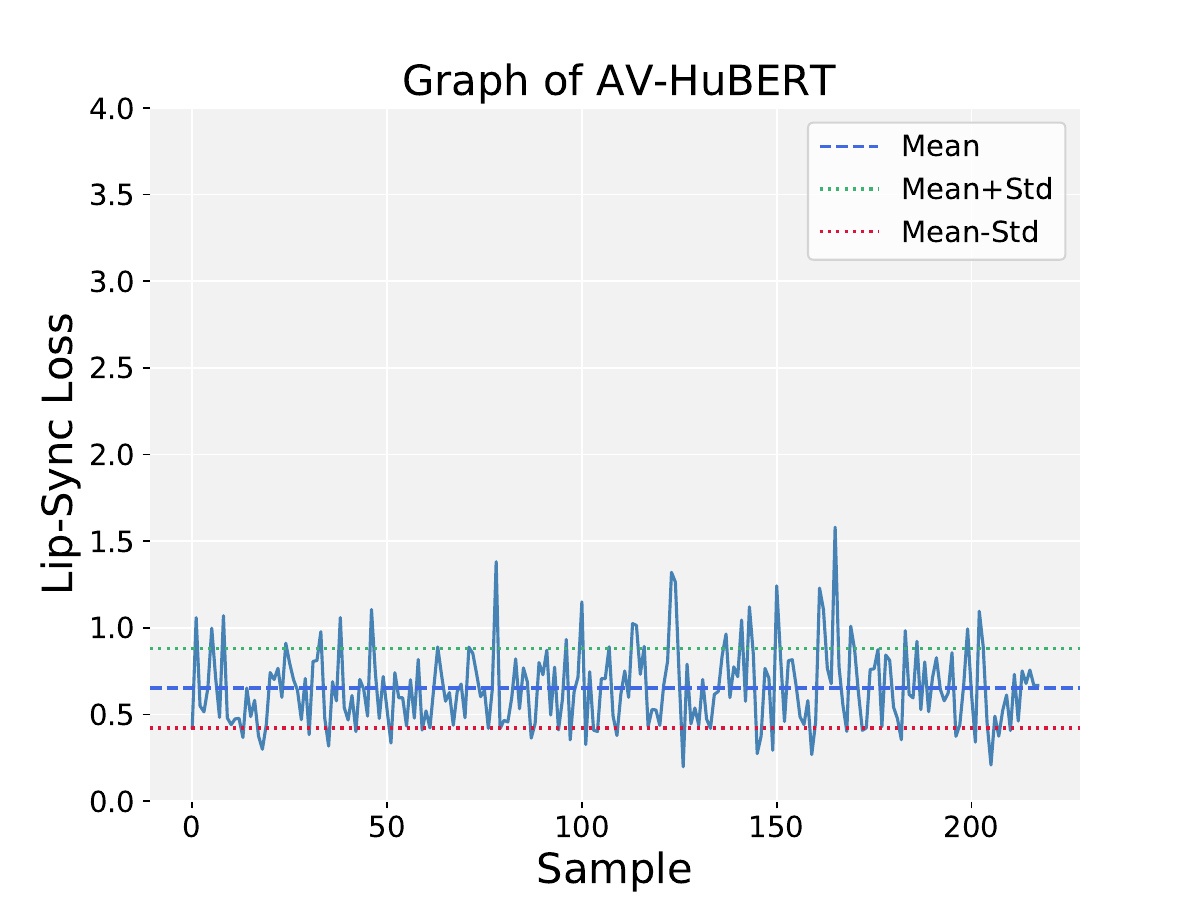}
    \caption{Lip-sync loss}
    \label{fig:logloss_avhubert}
  \end{subfigure}
  %\caption{\textbf{(a, b)} }
  %\label{fig:avhubert_results}
  \caption{Cosine similarity and lip-sync loss between GT audio-lip pairs on random LRS2 test samples, showcasing the instability of SyncNet~\cite{prajwal2020lip} \textbf{(a, b)} and more robust performance of AV-HuBERT \textbf{(c, d)}.}
  \label{fig:syncnet_analysis}
\end{figure}

In the talking face generation task, visual quality of the generated faces and audio-lip synchronization (lip sync) are essential but also challenging aspects to have a natural video. 
Since visual artifacts and out-of-sync lip movements are easily recognizable by the audience, they significantly diminish the naturalness of the dubbed video. 
While visual quality is addressed with approaches across various domains, lip sync takes precedence as it is particular and crucial for talking face generation in maintaining the naturalness of dubbed videos. 
Up to now, several different works tackled the challenges of learning lip sync. 
%The widely adopted method involves feature extraction from both the audio and face sequences, focusing specifically on the mouth region. 
The widely adopted method involves extracting features from both the audio and face sequences by a model that has been contrastively trained with audio-face pairs to learn lip sync.
These features are then compared to measure the synchronization between the two modalities (e.g., with cosine similarity). 
Therefore, it is essential to extract meaningful as well as robust audio and visual features, as the speech representations from both modalities significantly influence the synchronization measurement. 
%For this, the most common method in literature is using a \textit{lip-expert}, which is proposed by Wav2Lip~\cite{prajwal2020lip} inspired by SyncNet~\cite{chung2017out}. 
For this, the most common method is using a \textit{lip-expert}, which is a slightly modified and retrained version of SyncNet~\cite{chung2017out}\footnote{Therefore, SyncNet name is being used for the original SyncNet~\cite{chung2017out} and \textit{lip-expert}~\cite{prajwal2020lip} interchangeably, as we also do it hereinafter.}.
Specifically, the features from the lips and audio sequence are extracted by the SyncNet~\cite{prajwal2020lip} and the cosine similarity along with the cross-entropy loss is computed.
This loss strategy is called \textit{lip-sync loss}.
Significant advancements have been made since the introduction of using this SyncNet~\cite{prajwal2020lip} and most of the methods have benefited from this approach to learn lip sync in the literature.  
Recently, TalkLip~\cite{wang2023seeing} proposes using a lip-reading expert, AV-HuBERT~\cite{shi2022avhubert,shi2022avsr}, for audio and visual feature extraction. 
Then, contrastive learning is employed to learn lip sync, yielding superior performance. 
%Av-HuBERT lip-reading expert results in a more accurate lip sync performance.
%In a recent study, using a lip-reading expert for audio and visual feature extraction has been proposed, resulting in a more accurate lip synchronization performance. 

In ~\cref{fig:cosine_syncnet,fig:logloss_syncnet}, we share our analysis about the performance of SyncNet~\cite{prajwal2020lip} on LRS2~\cite{LRS2} ground-truth (GT) audio-lip pairs.
The cosine similarity and lip-sync loss show fluctuations even on GT samples, contrary to the anticipated stable and high performance.
This outcome implies that SyncNet~\cite{prajwal2020lip} has significant stability and reliability issues that lead to poor lip sync performance.
Moreover, we empirically find that using SyncNet~\cite{prajwal2020lip} causes severe visual quality issues and unstable training, despite enhanced lip sync performance.
%To tackle this problem, inspired by TalkLip, for audio and visual feature extraction for training our talking face generation model, we employ a pretrained audio-visual speech representation learning model, AV-HuBERT, that was finetuned for lip reading task.
To address these problems, inspired by TalkLip~\cite{wang2023seeing}, we employ a pretrained audio-visual speech representation learning model, AV-HuBERT~\cite{shi2022avhubert,shi2022avsr}, which was finetuned for the lip reading task, to extract audio and lip features. % for training our talking face generation model.
In contrast to TalkLip, we utilize cross-entropy-based lip-sync loss~\cite{prajwal2020lip} %to measure the lip sync %after feature extraction and employ this for cross-entropy-based loss 
to guide our model during training, providing a stabilized training signal (\cref{fig:cosine_sim_avhubert,fig:logloss_avhubert}).
We refer to this as approach as \textit{unsupervised}.
In addition to this, we investigate two further methods as loss functions.
Specifically, we employ AV-HuBERT features and compute the lip-sync loss by using the visual features of the generated faces and GT faces.
We call this approach \textit{visual-visual}, as the audio is not involved.
We also obtain features from generated face-audio pairs and GT face-audio pairs with AV-HuBERT model for lip-sync loss and term this \textit{multimodal}, since we acquire the features from the multimodal representation (face-audio pairs).
We conduct an ablation study about lip sync learning by comparing these introduced approaches and present the results in \cref{sec:eval:ablation}.
%Subsequently, we calculate the cosine similarity between these two feature representations to serve as a synchronization loss, guiding the network during the training.

Besides training a model with high-quality audio-lip synchronization, proper and robust evaluation of such capabilities is another key aspect of talking face generation, essential for analyzing and comparing different methods. 
One of the first lip sync evaluation metrics is Mouth Landmark Distance (LMD)~\cite{chen2018lip}, focusing on computing the distance between the landmarks in the mouth region of the generated faces and GT faces.
However, this metric does not disentangle the lip sync from visual factors, as it is sensitive against shifting in the spatial domain.
Additionally, lips with different articulatory parameters (e.g., aperture and spreading) yield poor LMD score, although they are still synchronized.
More recent metrics, LSE-C \& -D, are based on SyncNet~\cite{chung2017out} features and measure the confidence and distance scores to represent lip sync.
The advantage of these metrics is that they do not require GT data, directly measuring the alignment between audio and faces.
However, unstable SyncNet~\cite{chung2017out} performance makes these two metrics unreliable and also vulnerable to affine transformation.
One of the main reasons of this are the poor shift-invariant characteristics of SyncNet~\cite{chung2017out}.
To tackle these problems in lip sync evaluation, we propose three novel complementary metrics: 
Unsupervised Audio-Visual Synchronization (AVS$_u$), Multimodal Audio-Visual Synchronization (AVS$_m$), and Visual-only Lip Synchronization (AVS$_v$). 
We leverage the pretrained AV-HuBERT lip-reading expert for feature extraction and utilize cosine similarity for the score calculation.
The differences and details of these metrics are presented in \cref{sec:metrics}.
Our contributions are summarized as follows:

\begin{itemize}
    \item We propose to use a pretrained audio-visual speech representation learning model (AV-HuBERT), finetuned for lip reading task, for feature extraction from the audio and face sequences for lip-sync loss in training.
    \item We introduce three novel evaluation metrics by employing AV-HuBERT for feature extraction, yielding less vulnerable and more consistent assessments of performance.
    \item We conduct extensive experiments and ablation studies to demonstrate the effectiveness of our contributions. %We further validate our proposed metrics through a user study.
\end{itemize}

\section{Related Work}
\label{sec:related_work}

\paragraphHeading{Talking face generation}
Traditional methods focus on achieving time-aligned videos by choosing the most fitting image-audio pairs~\cite{yehia1998quantitative,brand1999voice,suwajanakorn2017synthesizing}.
Later on, the facial landmark representation for face generation is employed to obtain a synchronized lip~\cite{das2020speech,zhou2020makelttalk,chen2019hierarchical,zhong2023identity}. 

However, these methods suffer from poor lip sync, despite controllable face generation.
Wav2Lip~\cite{prajwal2020lip} proposes a \textit{lip-expert}, modified SyncNet~\cite{chung2017out}, and also a lip-sync loss to guide the model for lip sync learning.
It shows superior lip sync performance. % and achieves SOTA lip sync performance.
PC-AVS~\cite{zhou2021pose} introduces a pose-controllable 2D talking face generation without using any intermediate representation (e.g., facial landmarks, 3D head representation). % and also utilizes contrastive learning for lip sync.
GC-AVT~\cite{liang2022expressive} employs a similar approach as PC-AVS but involves emotion-controllable face generation.
SyncTalkFace~\cite{park2022synctalkface} benefits from an audio-lip memory mechanism to store and retrieve lip motion representation to achieve enhanced lip sync.
On the other hand, VideoReTalking~\cite{cheng2022videoretalking} reveals fundamental problems in talking face generation, namely the effect of lip motion of the identity reference over the talking face generation.
To solve this problem, they transform the identity reference to have canonical expression with stable and flat lips, which yields improved training stability and lip sync.
DINet~\cite{zhang2023dinet} employs a deformation module to improve the pose alignment and lip sync.
Recently, LipFormer~\cite{wang2023lipformer} introduced a pre-learned facial codebook-based method to learn HR video generation by overcoming existing challenges. %, as it is a challenging aspect of the talking face generation due to the lip sync loss and low-resolution datasets.
TalkLip~\cite{wang2023seeing} employs a global audio encoder to capture the content in the speech and also introduces AV-HuBERT-based audio-visual feature extraction for lip sync learning along with contrastive learning. 
However, TalkLip has severe visual artifacts, despite superior lip sync.
This method is the closest approach to our talking face generation. 
However, we use lip-sync loss instead of contrastive learning and investigate two further methods for lip sync learning. 
Moreover, we achieve better visual quality without artifacts.
On the contrary, in ~\cite{yaman2023audiodriven}, comprehensive analyses are provided regarding lip leakage observed in the reference image as well as the instability issues encountered with SyncNet. They propose a silent-lip generator to manipulate the reference image, mitigating lip leakage, and introduce stabilized synchronization loss to address the stability concerns within the SyncNet model and synchronization loss.
SIDGAN~\cite{muaz2023sidgan} presents crucial analyses of learning synchronization. 
They also provide a shift-invariant model, similar to the lip-expert \& SyncNet, for feature extraction to guide the model for lip sync learning.
On the other hand, one of the first attempts to generate synchronized lips as a part of a system was done in~\cite{ritter1999face} and the follow-up paper~\cite{waibel2023face} also contains an entire system that involves speech-to-speech translation and face dubbing. 
These systems are also well-suited for utilization in meeting rooms, facilitating seamless communication among individuals conversing in different languages~\cite{waibel2003smart}.
In addition to the above 2D-based approaches, Neural Radiance Fields-based (NeRFs) and 3D-based methods aim at synthesizing the entire head by representing the head in the 3D space~\cite{blanz1999morphable,booth2018large,zhou2019talking,zhou2020makelttalk,thies2020neural,wu2021imitating,zhang2021facial,zhang2021flow,yin2022styleheat,guo2021ad,yao2022dfa,liu2022semantic,shen2022learning,tang2022real,song2022everybody,papantoniou2022neural,ye2023geneface,wang2023progressive}. 
Though they are capable of controlling the pose and emotion much better as well as achieving enhanced visual quality, they have severe lip sync performance, yielding unrealistic videos.

\paragraphHeading{Lip sync evaluation}
%Lip synchronization evaluation is still an open research problem.
The very first proposed metric is Lip Landmark Distance (LMD)~\cite{chen2018lip}.
However, it has several issues, as we mentioned in \cref{sec:intro} and \cref{sec:metrics}.
After the SyncNet~\cite{chung2017out} and Sync scores~\cite{chung2017out} were proposed, they demonstrated a more convenient performance than LMD.
Later, Wav2Lip~\cite{prajwal2020lip} proposed LSE-C and LSE-D metrics (confidence and distance) using SyncNet~\cite{chung2017out} audio and visual features, which became the gold standard in the literature.
%These metrics are confidence and distance scores for lip synchronization.
Despite the advantage of not requiring GT data, the unreliable performance of SyncNet makes these two metrics vulnerable.
The Word Error Rate (WER) has recently been proposed as an evaluation metric for reading intelligibility~\cite{wang2023seeing}.
%The authors utilized a lip-reading expert to obtain the transcription of the generated video and then compare it with the GT transcript.
In this work, we propose three novel lip sync evaluation metrics by utilizing a robust pretrained audio-visual speech representation learning model, AV-HuBERT.

\begin{figure*}[th]
  \centering
   \includegraphics[width=0.93\linewidth]{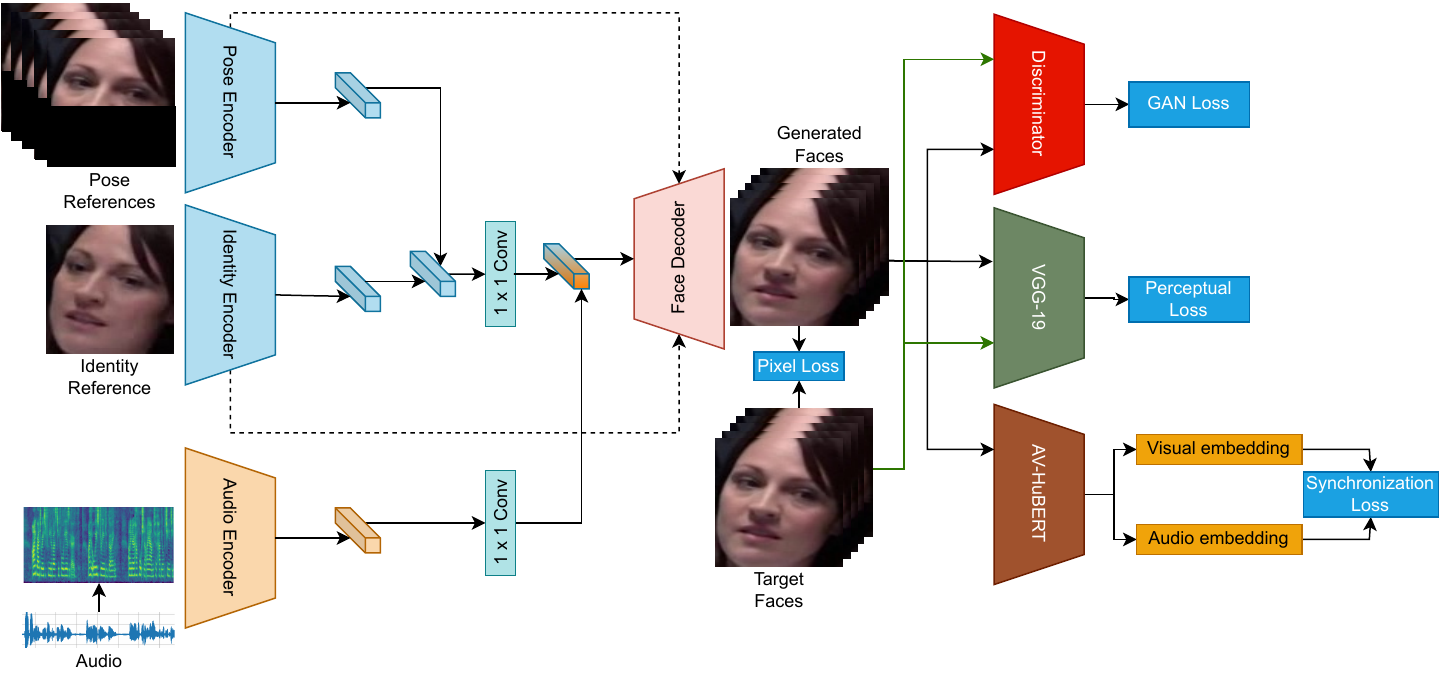}
   %\vspace{-0.6cm}
   \caption{Illustration of the proposed audio-driven talking face generation model and employed loss functions. }
   \label{fig:method}
\end{figure*}

\section{Talking Face Generation}

We propose a talking face generation approach to enhance lip sync and visual quality by leveraging an audio-visual lip-reading expert named AVHubert~\cite{shi2022avhubert,shi2022avsr}. 
\cref{fig:method} provides an overview of our model, which takes three inputs: an identity reference, a bottom-half masked pose reference, and an audio snippet. 
The face generator is responsible for synthesizing a set of images to retain synchronized lips with respect to the given audio while preserving the identity and visual quality. 
We extract features from the generated samples with the AV-HuBERT and then calculate lip-sync loss in the training along with the other losses.

\subsection{Face Encoder}

In 2D talking face generation, the main approach is to provide an identity reference and a pose reference to the model, which aims to synthesize a modified version of the pose reference with lip movements matching the given audio. 
Since the target image and the pose reference are the same, the mouth region of the pose reference has to be masked. 
On the other hand, the identity reference is utilized to preserve the subject's identity and is randomly sampled from a different part of the input video than the pose reference.
To encode the pose and identity references, we employ two individual encoders. 
This approach demonstrates superior performance than using a single encoder for both modalities as in the traditional approaches, since each encoder follows its own objective more effectively, yielding better visual feature representation for both inputs~\cite{muaz2023sidgan}. 
Our identity and pose encoders share the same architecture. 
They have consecutive convolutional blocks and each block involves one strided-convolution layer followed by two non-strided convolution layers. 
After each layer, we employ a ReLU activation function~\cite{nair2010rectified,krizhevsky2012imagenet} and a batch normalization layer~\cite{ioffe2015batch}. 

\subsection{Audio Encoder}

Our audio encoder embeds the mel-spectrogram representation of the audio snippet, acting as a condition for the face generator to drive the model to generate accurate lip movements, $ E_A(A) = F^A \in \mathbb{R}^{1 \times 1 \times 512} $. 
%We employ a pretrained audio encoder that was trained in conjunction with a face encoder to learn lip synchronization evaluation. 
We employ the audio encoder of the SyncNet~\cite{prajwal2020lip} without finetuning, since it was trained in conjunction with a face encoder to learn lip synchronization.
In this way, the audio encoder provides a feature representation more suitable for the purpose of generating synchronized lips. 

\subsection{Video Generation}

The talking face generator takes the combination of features of three encoders. 
We first concatenate the identity features and pose features along the depth dimension. 
Before feeding the face generator, we process these features through a convolution layer to reduce the depth. 
Subsequently, we concatenate the output with the audio embedding to input the face generator. 
Our face generator consists of consecutive transposed convolution layers. 
Following each transposed convolution layer, we utilize two convolution layers with a stride one. 
Similar to the image encoders, we employ the ReLU activation function~\cite{nair2010rectified,krizhevsky2012imagenet} and batch normalization~\cite{ioffe2015batch} after each layer. 
We also apply residual connections~\cite{he2016deep} between the reciprocal layers of the face encoders (both identity and pose encoders) and the face generator. 
This strategy yields the retention of high-level features and increased stabilization throughout the training.

\subsection{Audio-Visual Speech Representation Expert}

Audio-Visual Hidden Unit BERT (AV-HuBERT)~\cite{shi2022avhubert,shi2022avsr} is a self-supervised representation learning model for audio and visual data. 
The model processes a face sequence (the mouth region) with a modified version of ResNet-18 and the Mel-frequency cepstral coefficients (MFCC) of an audio sequence with a linear projection layer followed by a normalization step based on per-frame statistics. 
The fusion of audio and visual features is processed through transformer blocks to predict masked cluster assignments. 
Thus, AV-HuBERT learned robust audio-visual speech representation.
We utilize the finetuned version of AV-HuBERT for lip reading, since it yields better lip sync as well as reading intelligibility~\cite{wang2023seeing}.
In \cref{fig:syncnet_analysis}, we show the performance of AV-HuBERT on LRS2 GT data for measuring the cosine similarity and lip-sync loss.
The graphs clearly show that the AV-HuBERT has more stable performance and less fluctuation compared to SyncNet~\cite{prajwal2020lip} used in Wav2Lip~\cite{prajwal2020lip}. % (see \cref{fig:syncnet_analysis}).

\subsection{Lip Synchronization Loss}

\begin{figure}[t]
  \centering
   %\includegraphics[width=\linewidth]{sec/AVHuBERT_input_preparation.pdf}
   \includegraphics[width=\linewidth]{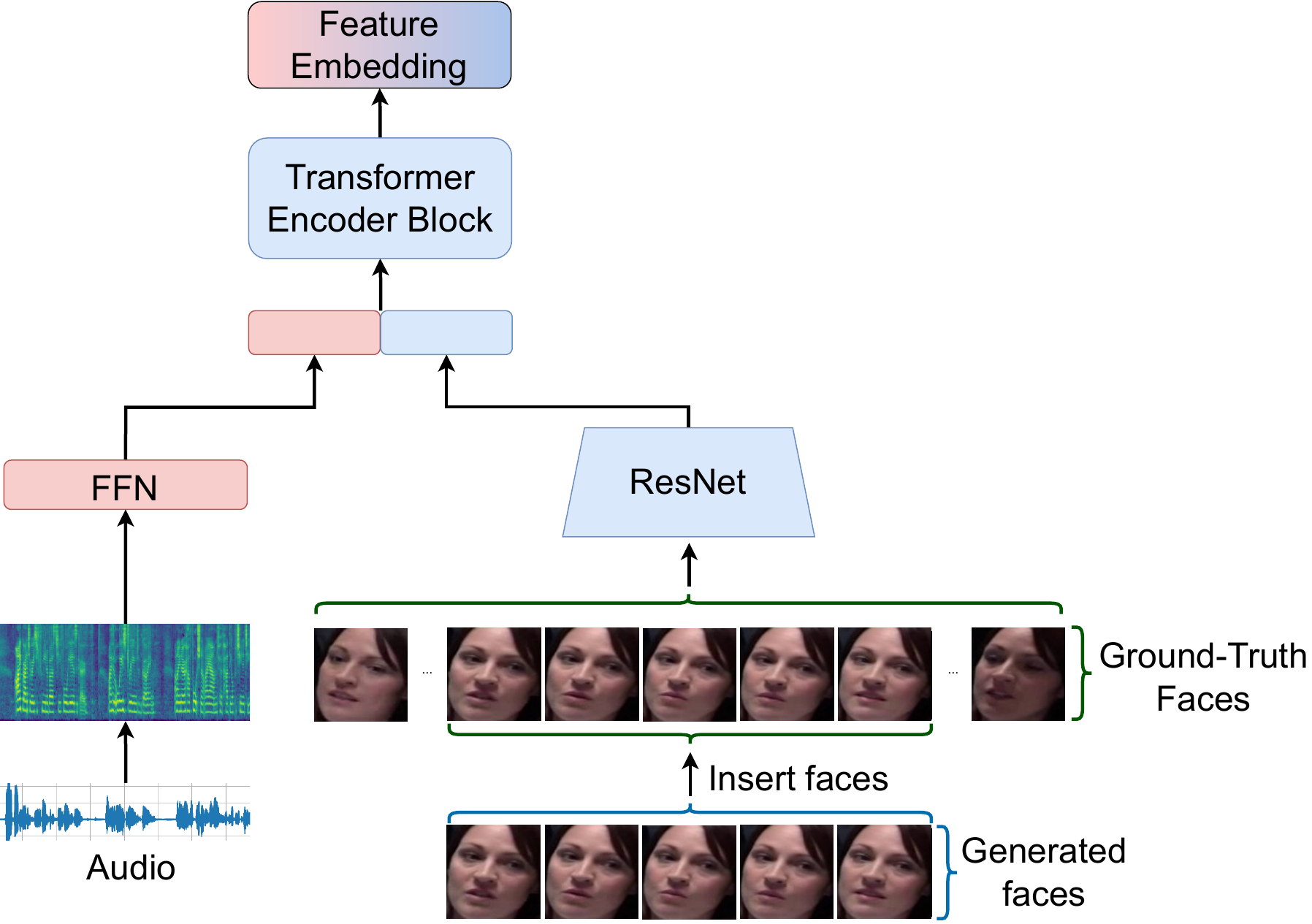}
   %\vspace{-0.5cm}
   \caption{In the training, since extracting features from entire videos provides more informative features~\cite{wang2023seeing}, we insert the generated faces into the corresponding part in the target video. Then, we use this video for feature extraction after cropping the mouth. }
   \label{fig:video_preparation}
\end{figure}

We utilize the pretrained AV-HuBERT model~\cite{shi2022avhubert,shi2022avsr} and extract features from the final layer of the transformer encoder block for audio and video modalities: $ F^A_{AVH} \in \mathbb{R}^{T \times 768}, F^V_{AVH} \in \mathbb{R}^{T \times 768}  $. 
Along with ~\cite{wang2023seeing}, we empirically found that extracting features from the entire video instead of a short sequence yields better audio-visual feature alignment. 
Therefore, we specifically replace the corresponding interval of the ground-truth video with the generated face sequence (\cref{fig:video_preparation}). 
We then extract features from the video using the cropped lips from the face sequence and audio sequence to calculate the lip sync between audio and visual features. 
However, since a major part of the video is ground-truth data, it is anticipated to have high audio-visual alignment, yielding insignificant effects of the generated samples on lip synchronization evaluation. 
To tackle this problem, we only consider the generated samples in the feature space for lip sync loss, as the AV-HuBERT feature extractor provides the feature representation per time step, $ T \times D $. 
Specifically, we take the audio and visual AV-HuBERT features corresponding to the generated time interval. 
Subsequently, we compute cosine similarity between these two feature representations followed by binary cross-entropy loss as follows: 
\begin{equation}
    L_{sync} = -log(CS(F^{A_{t:t+k}}_{AVH}, F^{V_{t:t+k}}_{AVH}))
\end{equation}
where $ CS $ indicates cosine similarity and $ t:t+k $ represents the time interval of the generated part of the video. 
$ k $ is the same as the length of the face sequence generated by the talking face generation model in a single forward pass.

\subsection{Implementation Details}

\paragraphHeading{Adversarial loss} 
We utilize the GAN (adversarial) loss~\cite{goodfellow2014generative} to train our talking face generation model. 
For this, we employ a discriminator model, which is responsible for distinguishing real (target data) and fake samples (generated data) to guide the generator. 
Meanwhile, the generator attempts to synthesize appropriate images so that the discriminator cannot determine whether the sample is real or fake. 
Our discriminator network contains $ 7 $ consecutive strided-convolutional layers along with Leaky ReLU activation function and spectral normalization~\cite{miyato2018spectral}.

\paragraphHeading{Perceptual loss} 
In order to preserve the identity and textures, we utilize perceptual loss~\cite{johnson2016perceptual} as feature reconstruction loss by extracting features from the generated faces and GT faces from different layers of the pretrained VGG-19 model~\cite{simonyan2014very}.
Afterward, we calculate the L2 distance between extracted features, as shown below. 
While $ c_i $ indicates weight coefficients, $ \phi $ states the selected layers for the feature extraction. 
$ I^G $ and $ I^{GT} $ are the generated image and the GT image, respectively. 
We follow \cite{johnson2016perceptual} for determining the coefficients and layers.
\begin{equation}
    L_{per} = \sum_{i=1}^{5} c_i ||VGG^{\phi_i}(I^G) - VGG^{\phi_i}(I^{GT})||_2
\end{equation}

\paragraphHeading{Pixel reconstruction loss} 
Despite perceptual loss to capture identity and textural details, pixel-level reconstruction loss is required to capture fine-grained details and generate consistent images. 
Therefore, we employ a reconstruction loss in the pixel space: $ L_{pixel} = ||I^G - I^{GT}||_1 $

\paragraphHeading{Total loss}
By combining all the presented loss functions, the total loss is as follows:
\begin{equation}
    L = L_{GAN}(G, D) + \lambda_1 L_{pixel}(G) + \lambda_2 L_{per}(G) + \lambda_3 L_{sync}
\end{equation}
where $G$ and $D$ denote the generator and discriminator. % outputs, respectively. 
We determined the best coefficients through empirical analysis as follows: $ (\lambda_1, \lambda_2, \lambda_3) = (10, 1, 0.5) $.

\paragraphHeading{Training details} 
In each forward pass, we generate a set of images (denoted by $k$) to ensure temporal consistency. 
Following the literature, we set $k = 5$.
We use FAN~\cite{bulat2017far} for face detection and obtain tight crops as input. 
We resize the faces to $96 \times 96$ resolution since lip sync learning in high resolution holds further challenges~\cite{muaz2023sidgan} and the LRS2 dataset~\cite{LRS2} has low-resolution faces. 
Then, we can apply GFPGAN~\cite{wang2021gfpgan} face enhancement method to the output video to increase the resolution for obtaining HR videos, if necessary. 
Our audio encoder takes a mel-spectrogram of size $16 \times 80$ extracted from $16 kHz$ audio. 
The hop and the window sizes are $200$ and $800$, respectively. 
We utilize the Adam optimizer~\cite{kingma2014adam}. 
The learning rate is set to $1 \times 10^{-4}$.

\definecolor{ourgreen}{HTML}{d2f0aa}
\definecolor{ouryellow}{HTML}{fcef9e}

\section{Experimental Results}
\label{sec:eval}

\paragraphHeading{Dataset}
We developed our talking face generator by utilizing the standard benchmark in the domain due to its diversity in terms of number of subjects: Lip Reading Sentence 2 (LRS2)~\cite{LRS2} training set. 
%The evaluation was conducted on both the LRS2 and LRW test sets without further finetuning the model on LRW.
The evaluation was conducted on the LRS2~\cite{LRS2}, LRW~\cite{LRW}, and HDTF~\cite{zhang2021flow} datasets.
%We further evaluated our model on HDTF dataset~\cite{zhang2021flow}.

\subsection{Metrics}
\label{sec:metrics}
For visual quality assessment, we utilize benchmark metrics in this field: FID~\cite{heusel2017gans}, SSIM~\cite{wang2004image}, and PSNR. 
In evaluating lip sync, as in the literature, we employ Mouth Landmark Distance (LMD)~\cite{chen2018lip} as well as LSE-C \& LSE-D metrics~\cite{prajwal2020lip}, which measure the confidence and distance scores via a pretrained SyncNet~\cite{chung2017out}. 
We also conduct a user study to perform human evaluation.
We compare our model with the state-of-the-art models that have publicly available codes and models, ensuring a fair comparison by evaluating under the same conditions.

The three widely utilized synchronization metrics in the literature have crucial problems. 
The mouth landmark distance (LMD) measures the spatial difference between the mouth landmark points in the generated face and the target face. 
However, this metric faces the following three issues: 
(1) It is sensitive to errors in landmark detection. 
(2) It cannot disentangle the synchronization and the generation stability. 
For instance, if the model accurately generates lip movements but introduces errors in the mouth region (such as shifting the mouth/face), the landmark positions will subsequently change. 
Although this does not affect the synchronization, it leads to a higher landmark distance, highlighting the need for disentanglement. 
(3) LMD fails to consider lip movements or shapes properly. 
Despite having the same lip shape, variation in the lips aperture and spreading cause misleading scores.

\begin{table*}[!t]
  \caption{Quantitative results on the test sets of LRS2 and LRW. While \colorbox{ourgreen}{green} indicates the best score, \colorbox{ouryellow}{yellow} shows the second best.} % \emph{}
  \label{tab:results}
  \centering
  \resizebox{\textwidth}{!}{\begin{tabular}{@{}l|ccccccccc | ccccccccc@{}}
    \toprule
    %Method & LRS2 & LRW\\
     & \multicolumn{9}{c}{\textbf{LRS2}} & \multicolumn{9}{c}{\textbf{LRW}} \\
    \midrule
    Method & SSIM $\uparrow$ & PSNR $\uparrow$ & FID $\downarrow$ & LMD $\downarrow$ & LSE-C $\uparrow$ & LSE-D $\downarrow$ & AVS$_u$ $\uparrow$ & AVS$_m$ $\uparrow$ & AVS$_v$ $\uparrow$ & SSIM $\uparrow$ & PSNR $\uparrow$ & FID $\downarrow$ & LMD $\downarrow$ & LSE-C $\uparrow$ & LSE-D $\downarrow$ & AVS$_u$ $\uparrow$ & AVS$_m$ $\uparrow$ & AVS$_v$ $\uparrow$ \\ %PSNR & SSIM & FID & LMD & LSE-C & LSE-D \\
    \hline
    Wav2Lip~\cite{prajwal2020lip} & 0.865 & 26.538 & 7.05 & 2.388 & 7.594 & 6.759 & 0.248 & 0.659 & 0.289 & 0.851 & 25.144 & 6.81 & 2.147 & \colorbox{ouryellow}{7.490} & 6.512 & 0.242 & 0.537 & 0.268 \\
    VRT~\cite{cheng2022videoretalking} & 0.841 & 25.584 & 9.28 & 2.612 & 7.499 & 6.824 & 0.361 & 0.763 & 0.426 & 0.873 & 27.110 & \colorbox{ourgreen}{5.30} & 2.390 & 6.598 & 7.123 & 0.383 & 0.758 & 0.557 \\
    DINet~\cite{zhang2023dinet} & 0.785 & 24.354 & \colorbox{ouryellow}{4.26} & 2.301 & 5.376 & 8.376 & 0.291 & 0.758 & 0.425 & 0.886 & 27.501 & 8.17 & 1.963 & 5.249 & 9.099 & 0.276 & 0.638 & 0.360 \\
    TalkLip~\cite{wang2023seeing} & 0.860 & 26.112 & 4.94 & 2.344 & \colorbox{ourgreen}{8.530} & \colorbox{ourgreen}{6.086} & \colorbox{ourgreen}{0.570} & \colorbox{ouryellow}{0.895} & \colorbox{ouryellow}{0.702} & 0.868 & 26.349 & 15.73 & 1.836 & 7.281 & \colorbox{ouryellow}{6.485} & \colorbox{ourgreen}{0.581} & \colorbox{ouryellow}{0.813} & \colorbox{ouryellow}{0.604} \\
    IPLAP~\cite{zhong2023identity} & \colorbox{ouryellow}{0.877} & \colorbox{ouryellow}{29.670} & \colorbox{ourgreen}{4.10} & \colorbox{ouryellow}{2.119} & 6.495 & 7.165 & 0.331 & 0.711 & 0.479 & \colorbox{ouryellow}{0.917} & \colorbox{ourgreen}{30.456} & 8.40 & \colorbox{ouryellow}{1.641} & 5.949 & 7.767 & 0.272 & 0.593 & 0.331 \\
    \hline
    Ours & \colorbox{ourgreen}{0.947} & \colorbox{ourgreen}{31.273} & 4.51 & \colorbox{ourgreen}{1.188} & \colorbox{ouryellow}{7.958} & \colorbox{ouryellow}{6.301} & \colorbox{ouryellow}{0.508} & \colorbox{ourgreen}{0.939} & \colorbox{ourgreen}{0.879} & \colorbox{ourgreen}{0.919} & \colorbox{ouryellow}{30.185} & \colorbox{ouryellow}{6.21} & \colorbox{ourgreen}{1.487} & \colorbox{ourgreen}{7.738} & \colorbox{ourgreen}{6.456} & \colorbox{ouryellow}{0.554} & \colorbox{ourgreen}{0.856} & \colorbox{ourgreen}{0.762} \\
  \bottomrule
  \end{tabular}}
\end{table*}

Recently proposed LSE-C and LSE-D metrics are more informative than LMD. 
However, they also hold vital issues. 
These metrics rely on audio and lip features extracted by the pretrained SyncNet model~\cite{chung2017out}, which contains audio and image encoders, and was trained with audio-lip pairs for learning lip sync. 
%However, SyncNet exhibits unstable predictions when tested with GT face-audio pairs. 
However, SyncNet is vulnerable against translations in the data due to not being properly shift invariant~\cite{muaz2023sidgan} (see \cref{fig:lsecd_vs_avsu}).
%due to convolutional layers with stride two for downsampling. 
Therefore, small translations in the face affects LSE-C and LSE-D metrics, resulting in not fully disentangled lip synchronization evaluation.
Moreover, the margin around the faces has also impact on extracted features by SyncNet, yielding inconsistent LSE-C \& D scores.
%are sensitive and not consistent in assessing the audio-visual synchronization. 
To tackle these problems, we introduce three novel lip synchronization evaluation metrics. 
We employ AV-HuBERT, a robust audio-visual speech representation learning model, to obtain superior audio and visual feature representations to measure synchronization.

% Unsupervised Audio-Visual Synchronization (UAVS)
\paragraphHeading{Unsupervised Audio-Visual Synchronization (AVS$_u$)} 
In this metric, we measure the lip sync by only considering the given audio and the generated video.
Specifically, we extract audio and visual features (lips) from the transformer encoder block of AV-HuBERT. 
Subsequently, we compute the cosine similarity between these two feature representations, as several works demonstrate the superiority of cosine similarity for audio-visual alignment~\cite{prajwal2020lip,wang2023seeing} and speaker recognition~\cite{chung2003defence} . 
This metric does not require GT data and thus can flexibly be applied to any data, as LSE-C \& D. 
As AV-HuBERT provides more robust feature representation than SyncNet~\cite{chung2017out}, this metric is more consistent, reliable, and not vulnerable to translation (see \cref{fig:syncnet_analysis} and \cref{fig:syncnet_vs_avhubert_shift_and_rotation}). %and domain changes (e.g., different datasets).
\begin{equation}
    %AVS_{u} = CS(F^{V_{1:T}}_{G}, F^{A_{1:T}}_{GT})
    \text{AVS}_{u} = CS(TE(V^G_{1:T}, \mathbf{0}), TE(\mathbf{0}, A_{1:T}))
\end{equation}
where $ CS $ indicates the cosine similarity and $ TE $ represents the transformer encoder block from which we extract features.
$V^G$~and $A$~state the lips of the generated faces and audio sequence, respectively.
Please note that we give lips and audio to the AV-HuBERT model individually as we need to extract separate features to calculate the similarity between them.
However, the audio-visual transformer encoder of AV-HuBERT requires both modalities together. 
Therefore, we follow the approach proposed by~\citet{shi2022avhubert} and provide a placeholder of zeros with equivalent dimensions.

% Multimodal Audio-Visual Synchronization (MAUS)
\paragraphHeading{Multimodal Audio-Visual Synchronization (AVS$_m$)} 
In this metric, we employ the generated and GT videos as pairs to measure the synchronization. 
We first extract features from the generated video using the mouth region of the faces and audio. 
We then repeat the same procedure with the GT video. 
In the end, we calculate the cosine similarity between these two embeddings. 
Intuitively, this metric considers the similarity between the alignment of the generated lips-audio and GT lips-audio pairs. 
%This method explores the visual similarity of the lip movements of the generated and GT videos while involving the audio sequence.
\begin{equation}
    \text{AVS}_m = CS(TE(V^G_{1:T}, A_{1:T}), TE(V^{GT}_{1:T}, A_{1:T}))
\end{equation}

% Visual-only Lip Synchronization (VoLS)
\paragraphHeading{Visual-only Lip Synchronization (AVS$_v$)} 
We only employ the lips of the generated faces and GT faces without involving audio. 
Therefore, this metric only focuses on the visual shape similarity of the lips. 
As AV-HuBERT is finetuned for lip-reading purpose, the extracted visual features hold the information for lip reading, yielding meaningful representation for lip synchronization and intelligibility.
\begin{equation}
    \text{AVS}_v = CS(TE(V^G_{1:T}, \mathbf{0}), TE(V^{GT}_{1:T}, \mathbf{0}))
\end{equation}

\subsection{Quantitative Evaluation}
We demonstrate quantitative results on the test data of LRS2 and LRW in \cref{tab:results} and HDTF in \cref{tab:HDTF_results}. 
On LRS2 and HDTF, we surpass all other methods in the visual quality, except FID on LRS2, where we are slightly behind IPLAP and DINet.
On LRW, we achieve similar visual quality scores with IPLAP for SSIM and PSNR, and with VRT for FID.
In summary, we surpass other methods with a large margin in most of the visual quality metrics. 

\begin{figure}[tb]
  \centering
  \begin{subfigure}{0.49\linewidth}
    \includegraphics[trim={0cm 0cm 0cm 0.6cm},clip,width=\linewidth]{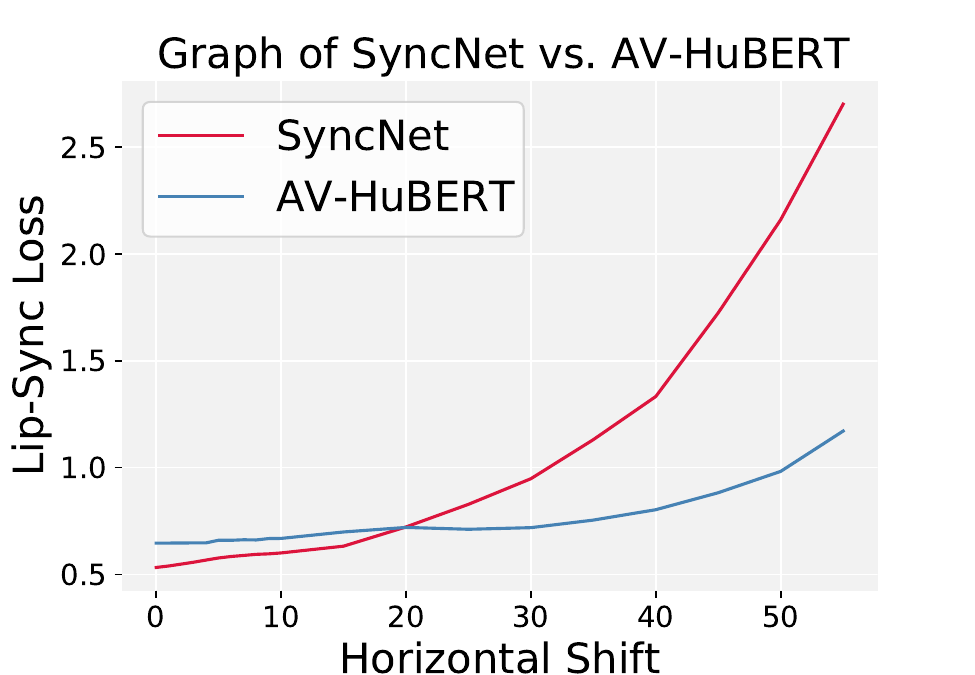}
    \caption{Horizontal shift analysis}
    \label{fig:logloss_horizontal}
  \end{subfigure}
  \hfill
  \begin{subfigure}{0.49\linewidth}
    \includegraphics[trim={0cm 0cm 0cm 0.6cm},clip,width=\linewidth]{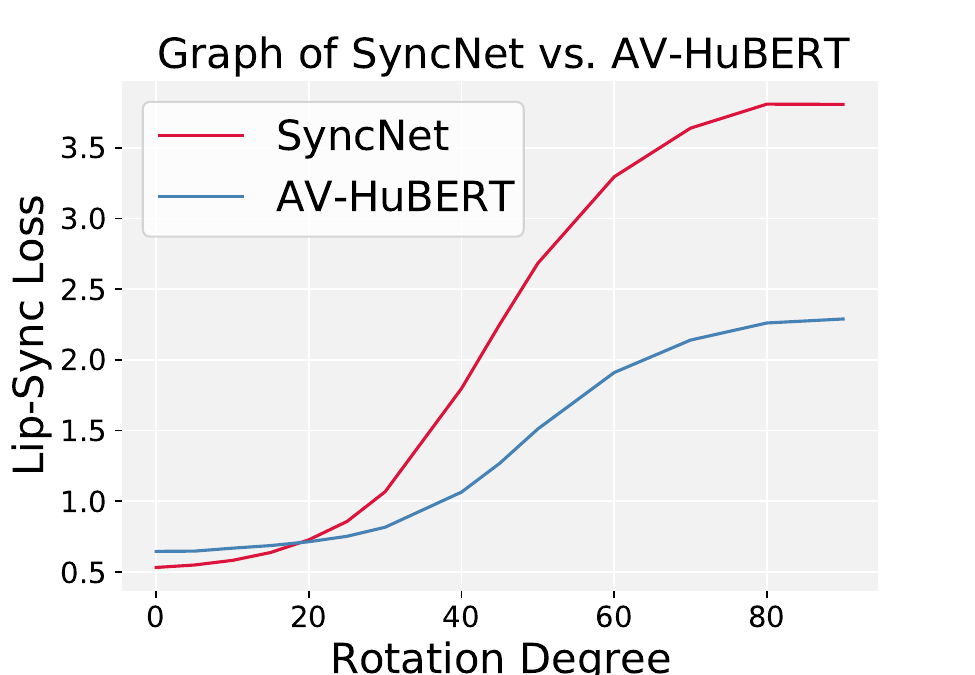}
    \caption{Rotation analysis}
    \label{fig:logloss_rotation}
  \end{subfigure}
  \begin{subfigure}{0.54\linewidth}
    \includegraphics[trim={0cm 0cm 0cm 0.7cm},clip,width=\linewidth]{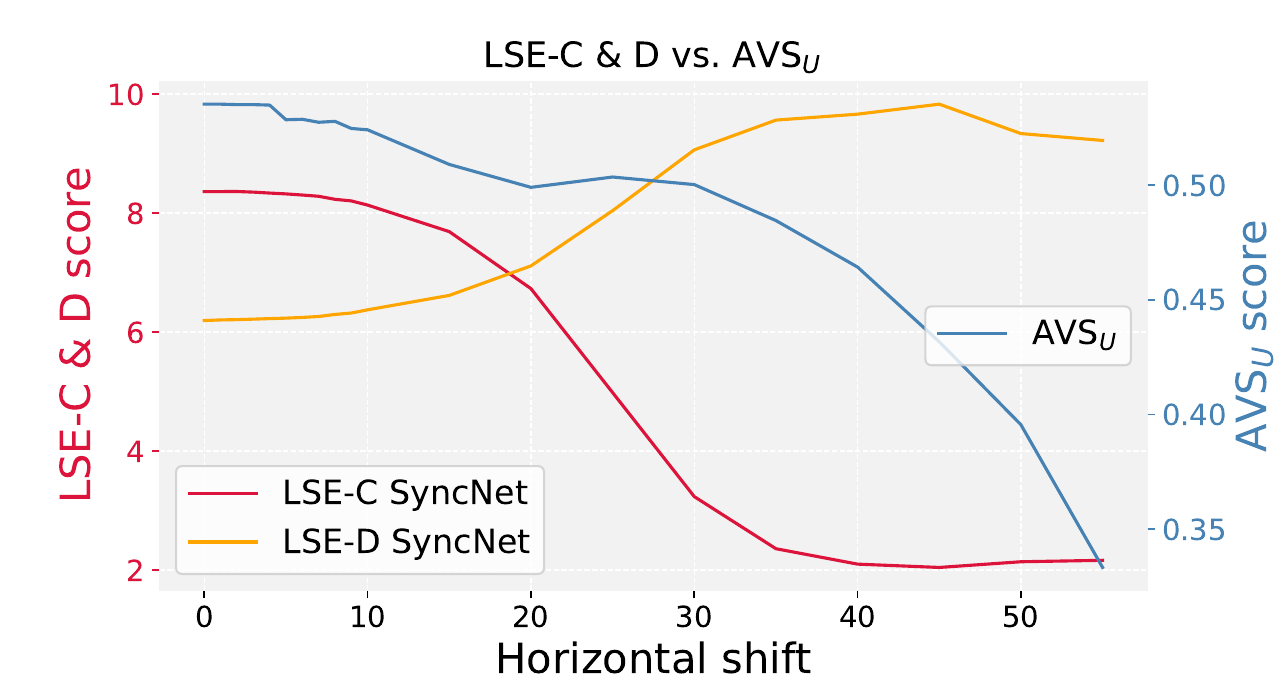}
    %logloss_vanilla_syncnet_vs_avhubert_categories
    \caption{Comparison of evaluation metrics}
    \label{fig:lsecd_vs_avsu}
  \end{subfigure}
  \hfill
  \begin{subfigure}{0.45\linewidth}
    \includegraphics[trim={0cm 0cm 0cm 0cm},clip,width=0.93\linewidth]{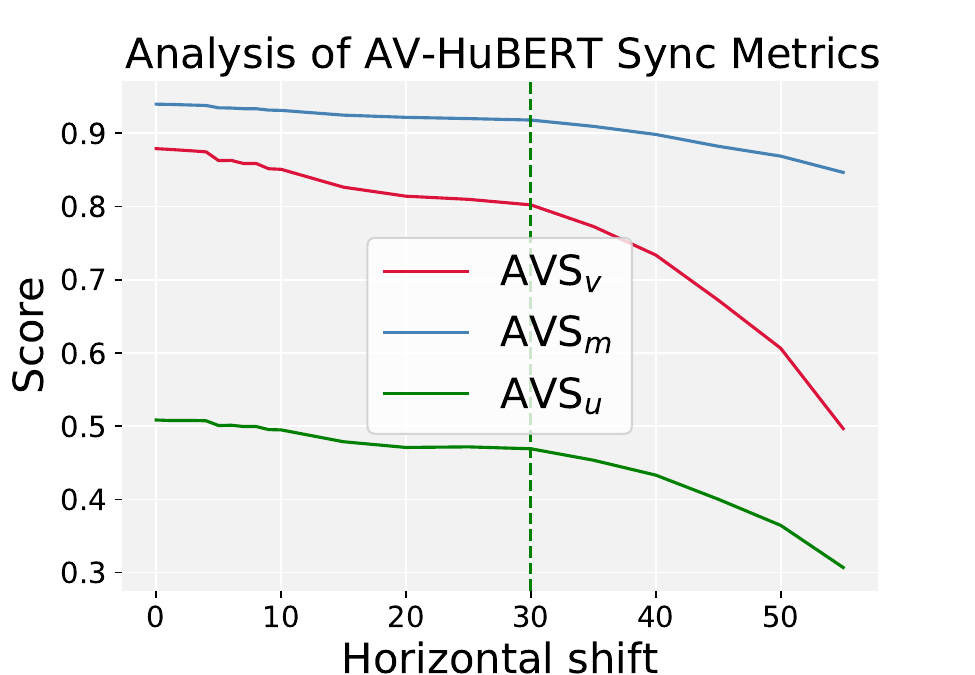}
    \caption{Analyses of our metrics}
    \label{fig:our_metrics}
  \end{subfigure}
  \caption{\textbf{(a,b)} shows the performance analyses of SyncNet~\cite{prajwal2020lip} and AV-HuBERT features for lip-sync loss on GT LRS2 data with the horizontal shift and rotation in spatial space. It clearly shows that SyncNet~\cite{prajwal2020lip} is not shift invariant and vulnerable to the affine transformation, while AV-HuBERT demonstrates robust performance. \textbf{(c)} compares the LSE-C \& D metrics with our $AVS_u$ metric while applying horizontal shifting in the spatial space. While LSE-C \& D scores are aligned with the left axis, $AVS_u$ is alifned with the right one. \textbf{(d)} analyses our three metrics under the shifting conditions. Since $AVS_v$ and $AVS_m$ require generated data-GT pairs, we use GT LRS2 data and our model's output. On the other hand, for \textbf{(a,b,c)}, we only use GT LRS2 data.
  }
  \label{fig:syncnet_vs_avhubert_shift_and_rotation}
\end{figure}

In LMD, we achieve state-of-the-art results on all three datasets. 
This outcome does not only show the synchronization performance but also demonstrate the visual stability of our generated outputs. 
In LSE-C and LSE-D metrics, we have state-of-the-art performance on LRW dataset. 
While TalkLip has better scores on LRS2, Wav2Lip achieves the highest score on HDTF. 
However, when considering our user study (\cref{sec:eval:user_study}), Wav2Lip is far behind our method on HDTF in synchronization performance, as in LRS2 and LRW datasets.
This shows the vulnerability and inconsistency of LSE-C \& D metrics.
\cref{fig:lsecd_vs_avsu} illustrates the sensitive performance of LSE-C \& D metrics when the visual data is horizontally shifted~\cite{muaz2023sidgan}.
It clearly proves that these metrics are extremely sensitive to translation in the data, measuring poor lip synchronization performance when the face is shifted while preserving the same lip shape.
Moreover, SyncNet~\cite{chung2017out} demonstrates similar fluctuation on GT data as in \cref{fig:cosine_syncnet}.
All these analyses and outcomes validate the motivation of our proposed new lip synchronization metrics.  

\begin{figure*}[t]
  \centering
   \includegraphics[width=\linewidth]{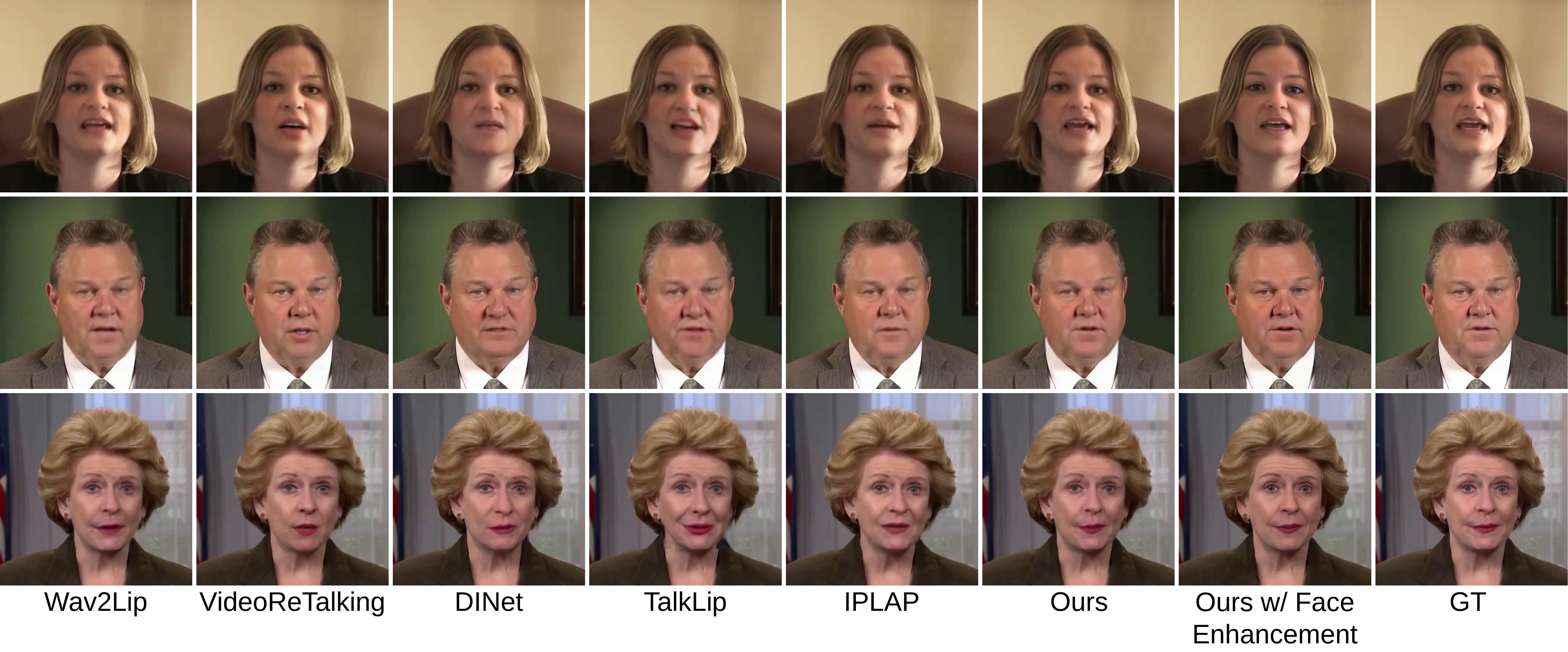}
   %\vspace{-0.7cm}
   \caption{Qualitative comparison of our approach with state-of-the-art models and ground-truth data on HDTF}
   \label{fig:sample_images}
\end{figure*}

According to our AV-HuBERT-based lip sync metrics, we achieve state-of-the-art results on all three datasets in AVS$_m$ and AVS$_v$. 
On the other hand, TalkLip surpasses us with a small difference on all three datasets in the AVS$_u$ metric.
In \cref{fig:cosine_sim_avhubert}, we illustrate the analysis of cosine similarity between AV-HuBERT lip and audio features on LRS2 GT data (similar to \cref{fig:cosine_syncnet}).
The graph shows that the AV-HuBERT features are more stable than SyncNet features.
Similarly, \cref{fig:lsecd_vs_avsu} and \cref{fig:our_metrics} also confirm the stability of our three novel metrics compared to LSE-C \& D. 
Considering these analyses as well as the results of our user study, it is clear that our proposed lip synchronization evaluation metrics provide more insight about the lip synchronization and also show more reliable performance.

\begin{table}[tb]
  \caption{Quantitative results on HDTF~\cite{zhang2021flow}.}
  \label{tab:HDTF_results}
  \centering
  \resizebox{\columnwidth}{!}{\begin{tabular}{@{}l|ccccccccc@{}}
    \toprule
    Method & SSIM $\uparrow$ & PSNR $\uparrow$ & FID $\downarrow$ & LMD $\downarrow$ & LSE-C $\uparrow$ & LSE-D $\downarrow$ & AVS$_u$ $\uparrow$ & AVS$_m$ $\uparrow$ & AVS$_v$ $\uparrow$ \\
    \midrule
    Wav2Lip & 0.841 & 24.812 & 35.41 & \colorbox{ouryellow}{1.341} & \colorbox{ourgreen}{9.054} & \colorbox{ourgreen}{6.414} & 0.297 & 0.514 & 0.358 \\
    VideoReTalking  & 0.830 & 24.551 & 29.77 & 3.085 & 6.121 & 7.368 & 0.384 & 0.677 & 0.570 \\
    TalkLip & 0.820 & 25.229 & 25.10 & 2.981 & 6.189 & 7.276 & \colorbox{ourgreen}{0.591} & \colorbox{ouryellow}{0.823} & \colorbox{ouryellow}{0.730} \\
    IPLAP & \colorbox{ouryellow}{0.869} & \colorbox{ouryellow}{27.801} & \colorbox{ouryellow}{22.09} & 2.217 & 5.563 & 8.495 & 0.459 & 0.661 & 0.528 \\
    Ours & \colorbox{ourgreen}{0.933} & \colorbox{ourgreen}{30.579} & \colorbox{ourgreen}{16.76} & \colorbox{ourgreen}{1.292} & \colorbox{ouryellow}{8.106} & \colorbox{ouryellow}{6.765} & \colorbox{ouryellow}{0.538} & \colorbox{ourgreen}{0.892} & \colorbox{ourgreen}{0.783} \\
  \bottomrule
  \end{tabular}}
\end{table}

\subsection{Qualitative Evaluation}
In \cref{fig:sample_images}, we present a qualitative comparison with SOTA models. 
We employ the respective publicly available models of the compared methods while generating videos. 
We choose the HDTF dataset to present results from an unseen dataset, except for DINet since it was trained on the HDTF dataset.
The results clearly show that our model surpasses all other methods in terms of having the most similar lip shapes with the GT face. 
Besides, mouth region and teeth have lower quality in Wav2Lip. 
Moreover, TalkLip has a severe pose stability issues and shows artifacts around the face. 
This clearly degrades the naturalness of a video.
Our user study in \cref{tab:user_study} validates the poor visual quality of TalkLip, DINet, and Wav2Lip.
In summary, the qualitative analysis and user study demonstrate the superiority of our model in terms of the lip synchronization and visual quality, resulting in natural talking face generation.
Moreover, we present results in \cref{fig:sample_images} by employing a face enhancement model, GFPGAN, to show that our model's results can be improved with a post-processing step to produce high resolution faces for high-resolution talking face video generation. 

\subsection{User Study}
\label{sec:eval:user_study}
We conduct a user study to explore how the generated videos look to humans.
We randomly select ten videos from the HDTF dataset and generate these videos with each model to use in the user study along with the GT videos.
Users were shown aligned videos of multiple faces generated by different methods and asked to rank them on lip synchronization, visual quality and overall quality.
In total, ten different participants joined the user study and we present the results in \cref{tab:user_study}.
The scores are scaled between $1$ (worst) and $5$ (best).
According to the results, we outperform all other models in lip synchronization, visual quality, and overall quality. 
TalkLip demonstrates the second-best synchronization performance.
However, its visual quality issues and artifacts yield the lowest visual quality score.

\begin{table}[t]
  \caption{User study on randomly selected HDTF videos~\cite{zhang2021flow}.} 
  \label{tab:user_study}
  \centering
  \footnotesize
  \begin{tabular}{@{}l|ccc@{}}
    \toprule
    Method & Sync $\uparrow$ & Visual $\uparrow$ & Overall $\uparrow$  \\ 
    \midrule
    Wav2Lip~\cite{prajwal2020lip} & 2.91 & 2.88 & 2.73 \\
    VideoReTalking w/ FR~\cite{cheng2022videoretalking} & 3.05 & 3.70 & 3.46  \\
    DINet~\cite{zhang2023dinet} & 2.50 & 2.34 & 2.48 \\
    TalkLip~\cite{wang2023seeing} & 3.32 & 2.05 & 2.08 \\
    IPLAP~\cite{zhong2023identity} & 2.62 & 3.86 & 3.27 \\
    Ours & \textbf{3.92} & \textbf{4.02} & \textbf{3.95}  \\
  \bottomrule
  \end{tabular}
\end{table}

\subsection{Ablation Study}
\label{sec:eval:ablation}
%\paragraphHeading{AV-HuBERT-based synchronization loss}

\begin{table}[tb]
  \caption{Ablation study on LRS2 dataset for AV-HuBERT-based synchronization losses.}
  \label{tab:ablation}
  \centering
  \resizebox{\columnwidth}{!}{\begin{tabular}{@{}l|ccccccccc@{}}
    \toprule
    Method & SSIM $\uparrow$ & PSNR $\uparrow$ & FID $\downarrow$ & LMD $\downarrow$ & LSE-C $\uparrow$ & LSE-D $\downarrow$ & AVS$_u$ $\uparrow$ & AVS$_m$ $\uparrow$ & AVS$_v$ $\uparrow$ \\
    \midrule
    Baseline & 0.864 & 26.424 & 12.25 & 2.423 & 7.116 & 7.396 & 0.301 & 0.637 & 0.423 \\
    Visual-visual & 0.905 & 29.248 & 15.14 & 1.798 & 7.481 & 6.556 & 0.381 & 0.765 & 0.545 \\
    Multimodal  & 0.910 & 30.014 & 6.11 & 1.774 & 6.998 & 6.794 & 0.395 & 0.789 & 0.575 \\
    Unsupervised & \textbf{0.947} & \textbf{31.273} & \textbf{4.51} & \textbf{1.188} & \textbf{7.958} & \textbf{6.301} & \textbf{0.508} & \textbf{0.939} & \textbf{0.879} \\
  \bottomrule
  \end{tabular}}
\end{table}

We conduct an ablation study to show the effect of using AV-HuBERT features in lip-sync loss throughout training.
For this, we first train our model with \textit{lip-expert}~\cite{prajwal2020lip} features for calculating lip-sync loss.
We call this model \textit{baseline} in \cref{tab:ablation}.
\textit{Unsupervised} represents the method that we propose to use.
Specifically, we extract features from the audio and lip sequences using the AV-HuBERT transformer encoder.
Afterward, we calculate lip-sync loss to explicitly measure the synchronization performance of the model in the training.
We also employ AV-HuBERT in two more ways to compare with our approach, following the AVS$_v$ and AVS$_m$ metrics.
In the \textit{visual-visual} approach, we extract only visual features by feeding the AV-HuBERT model with the generated lips and GT lips, individually.
Then, we apply lip-sync loss between these two embeddings without involving audio. 
This method obtains better scores than \textit{baseline}.
We further apply the \textit{multimodal} strategy, extracting features from generated lips-audio pairs and also from GT lips-audio pairs, and then applying lip-sync loss thereupon.
The \textit{multimodal} approach enhances the visual quality compared to \textit{baseline} and \textit{visual-visual} approaches but decreases the LSE-C \& D scores even below the \textit{baseline}. 
On the other hand, in our synchronization metrics, \textit{multimodal} outperforms the \textit{baseline} as well as \textit{visual-visual} method.
Finally, the best results in visual quality and lip synchronization are achieved by employing the \textit{unsupervised} approach according to the ablation study.

\begin{figure}[t]
  \centering
   \includegraphics[width=\linewidth]{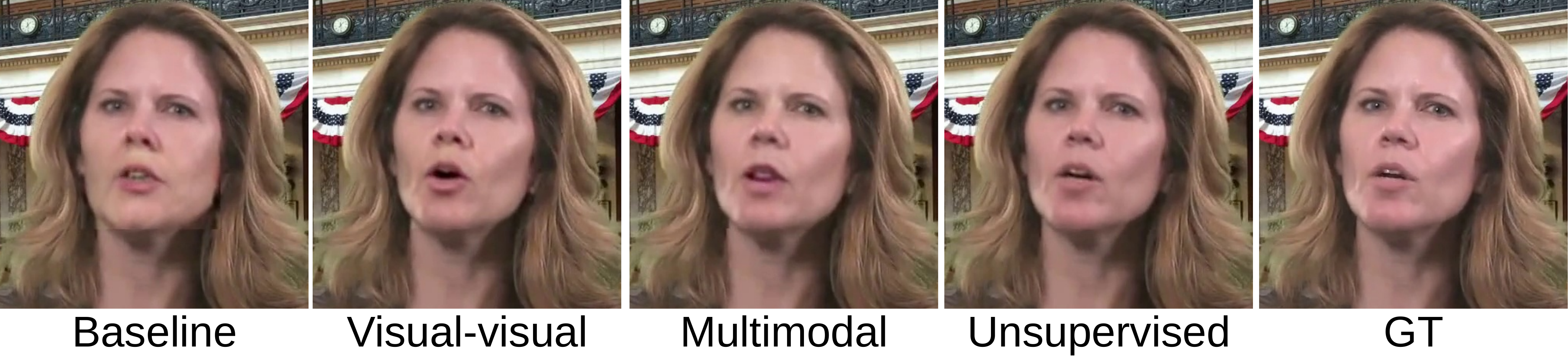}
   %\vspace{-0.65cm}
   \caption{Qualitative samples from the ablation study}
   \label{fig:ablation_study_faces}
\end{figure}

In \cref{fig:ablation_study_faces}, we share the sample images from the approaches presented in \cref{tab:ablation}. 
The \textit{baseline} has distinguishable face borders, artifacts in the mouth region, and not fully aligned lip shape.
Although \textit{visual-visual} shows enhanced visual quality and improved lip shape, \textit{multimodal} approach generates more appropriate lip shapes. 
Finally, \textit{unsupervised} is able to generate the best fitting lip shapes as well as enhanced visual quality. 
Specifically, the teeth have better visual quality and are more similar to the GT in terms of the characteristic features of the subject (identity).

\section{Conclusion}
\label{sec:conclusion}
We propose to use the pretrained audio-visual speech representation expert AV-HuBERT for training a talking face generation network with high-quality audio-lip synchronization.
Furthermore, we utilize this network to obtain three complementary and robust metrics for evaluating lip synchronization.
Our experimental results demonstrate the effectiveness of our approach.
We also analyze the proposed metrics for robustness and validate their alignment with human preferences through a user study.

\paragraphHeading{Limitations} 
AV-HuBERT and its features should be investigated further to employ them more efficiently for lip synchronization, despite increased performance and stability in the training.
Furthermore, the sample size of our user study could be increased to gain more statistical power.

\paragraphHeading{Ethics \& Social Impact}
Talking face generation is essential for a wide range of applications. 
However, its vulnerability and potential for misuse (e.g., deepfake) pose significant risks. 
We will apply Watermarking and take necessary precautions to prevent unauthorized usage of our model. 

\paragraphHeading{Acknowledgement} 
This work was supported in part by the European Commission project Meetween (101135798) under the call HORIZON-CL4-2023-HUMAN-01-03.

{
    \small
    \bibliographystyle{ieeenat_fullname}
    \bibliography{main}
}

\end{document}